\documentclass[runningheads]{llncs}
\usepackage{graphicx}
\usepackage{amsmath,amsfonts,amssymb,cancel,siunitx, latexsym}
\usepackage{algorithm}
\usepackage{algorithmicx}
\begin{document}
\title{Binary Orthogonal Non-negative Matrix Factorization }
%
%
\author{S. Fathi Hafshejani\orcidID{0000-0002-5731-8234} \and
D. Gaur\orcidID{0000-0001-6876-6000} \and
S. Hossain  \orcidID{0000-0003-1380-6241
} \and  R. Benkoczi  \orcidID{0000-0002-7942-0539}}
\authorrunning{S. Fathi Hafshejani et al.}
%
\institute{Department of Math and Computer Science,
    University of Lethbridge\\ 
    Lethbridge, AB, Canada\\
\email{sajad.fathihafshejan, daya.gaur, shahadat.hossain,robert.benkoczi@uleth.ca}}
\maketitle              
\begin{abstract}
We propose a method for computing binary orthogonal non-negative matrix factorization (BONMF) for clustering and classification. The method is tested on several representative real-world data sets. The numerical results confirm that the method has improved accuracy compared to the related techniques. The proposed method is fast for training and classification and space efficient. 

\end{abstract}
\section{Introduction}

For a data matrix $\bf X$ of size $m\times n$,
$\bf WH \approx X$ (where $\bf W$ is of size $m \times k$, $\bf H$ is of size $k \times n$) is considered as a low rank approximation ($k \ll n$) of the data matrix $\bf X$. Low rank approximations are essential in machine learning applications and especially in natural language processing and topic modelling where the data matrix is constructed over a collection of words from a vocabulary and a usually large collection of documents \cite{pauca2004text,shahnaz2006document,haddock2021semi,ding}. 

Singular value decomposition (SVD) \cite{wold1987principal} is an early approach for computing such a low rank approximation of data. SVD minimizes the Frobenius norm and the spectral norm simultaneously; not only that, the columns of $\bf W$ are orthogonal, and the rows of $\bf H$ are also orthogonal. However, the entries in $\bf W, H$ may be negative, which reduces the utility of SVD for data matrix $\bf X$ in which the entries are positive as the factors in $\bf W$ do not have an intuitive explanation. Non-negative matrix factorization (NMF), $\bf WH \approx X$ and $\bf X, W,H  \ge 0,$ was introduced by Paatero and Tapper \cite{paatero1994positive} to overcome this difficulty of interpretation of the factors. NMF was shown to be NP-complete by  Vavasis \cite{vavasis10}. NMF does not require the columns of $\bf W$ to be orthogonal, and this is considered a severe drawback in some applications as the columns (factors) of $\bf W$ are not separable by a large angle. Keeping this limitation in mind Ding et al. \cite{ding} introduced orthogonality constraints in NMF, $\bf X \approx WH$ and $\bf X, W, H \ge 0,$ the rows of $\bf H$ are orthogonal and demonstrated that is an effective approach for clustering of documents. 

We consider the following problem: given a $m \times n$ data matrix $\bf X$, we wish to represent $\bf X$ as a product of two matrices $\bf W, H$  with dimensions $m \times k$ and $k \times n$
respectively with the following restrictions: entries in $\bf W$ are positive, the entries in $\bf H$ are either $0$ or $1$, 
${\bf{ H H}}^T = I$ 
and the norm $|| {\bf{ X - WH }}||^2$ is minimized. Additionally, we want $k$ to be small compared to $n,m$. The columns of the data matrix $\bf X$ can be thought of as the $n$ samples. Low rank $\bf W$ represents the latent features. 
We call this problem the binary orthogonal nonnegative matrix factorization problem (BONMF).

\subsection{Contributions}

This paper gives a new method (Algorithm \ref{alg:ge}) for computing a binary orthogonal NMF using the two-phase iterative approach. In the first phase, we use a known update rule \cite{lee2009semi} to compute the factor $\bf W$. In the second phase, we use the observation that the binary constraints on $\bf H$ have a geometric interpretation. This gives an efficient rule to update $\bf H$ in each iteration (Equation (\ref{updat_h})). The entries in $\bf H$ are binary, and they are computed column-wise. If all the entries in $\bf H$ are non-zero, then $O(nk)$ space is needed. However, $\bf H$ is binary, and the rows of $\bf H$ are orthogonal. Therefore, only $O(n)$ space is needed. If we compute the entries of $\bf H$ columns-wise, intermediate states also need $O(n)$ space. The computation for each column of $\bf H$ takes $O(n^2k)$ steps. Therefore, the method is space efficient.

We evaluate the method's performance (in Section \ref{sec:empirical}) for training and testing on reference data sets from the ML repository. The experiments demonstrate that the training and the classification phase are efficient (Table \ref{experiments}). The method is accurate and outperforms the state of art methods (Table \ref{experiments}).
This method uses $k$ dot products of $m$ element vectors to update each column of the coefficient matrix $\bf H$ where $k$ is the number of classes in the data set. This is a significant reduction in the computation needed 
compared to the algorithms of \cite{lees,zhang2021non,ding} in the classification phase. The method is also space efficient as $\bf H$ is sparse.


\section{Related Work}
We begin with NMF and the related background needed to describe our algorithm. Given a non-negative matrix ${\bf {X}}\in\Bbb{R}^{m\times n}$, a non-negative matrix factorization of $\bf X$ finds two non-negative matrices ${\bf{ W}}\in \Bbb{R}^{m\times k}$ and ${\bf {H}}\in \Bbb{R}^{k\times n}$ with $k\ll \min(m,n)$ such that:

\begin{equation*}
    \bf X\approx WH,
\end{equation*}

and the entries in $\bf W, H$ are positive.
The factorization has a natural interpretation \cite{lees} and can be computed using various unsupervised machine learning methods. Due to its intuitive interpretation, NMF has found numerous applications such as data consolidation \cite{gao2019}, image clustering \cite{fu2019}, topic modelling \cite{arora2013practical}, community detection \cite{ye2021cdcn}, recommender systems \cite{ran2022differentially}, and gene expression profiling \cite{zhu2021ensemble}. 

BONMF is different from NMF. The entries in $\bf H$ are restricted to binary. If the columns of $\bf H$ are orthogonal, then the columns can be used to cluster the data. Therefore, BONMF factorization has several exciting applications \cite{zdunek2008data}.

Orthogonal NMF (ONMF) in which $\bf X \approx WH$ and $\bf W, H \ge 0$ and ${\bf{ H H}}^T = I$ was defined  by Ding et al. \cite{ding} who gave an algorithm based on solving the Lagrangian relaxation. 
The entries in $\bf  H$ in ONMF are not required to be binary. 
ONMF  use for data clustering was popularized by Seung and Lee \cite{seung2001algorithms}. One of the first notable applications of orthogonal NMF to document clustering is in \cite{yoo2010nonnegative} who gave improved algorithms and showed that ONMF performed better at document clustering than NMF. K-means \cite{macqueen1967classification} is one of the most widely used algorithms for unsupervised learning. Bauckhage
\cite{bauckhage2015k} showed that the objective function of K-means can be rewritten as ONMF if the entries in $\bf H$ are binary, and  the following condition holds: 

\begin{equation}
\label{eq:rcond}
\sum_i {\bf {H}}_{ij} = 1 \quad \forall j
\end{equation}

Therefore, BONMF is equivalent to K-means clustering. BONMF was also studied by Zdunek \cite{zdunek2008data} and differs from the well-studied non-negative matrix factorization (NMF). 
Lee et al. \cite{lee2009semi} studied BONMF without the condition \eqref{eq:rcond} on $\bf  H$ and gave an algorithm for determining such a factorization. However, applications to classification are not many. In this paper, we study BONMF for its use in prediction and classifying data, including clustering.
 


\subsection{NMF} 

NMF can be formulated as the following optimization problem that minimizes the square of the Frobenius norm:\footnote{$|| {\bf{A}}||_F = \sqrt{tr( {\bf{ A^T \times A}})} = \sum_{i,j} |a_{ij}|^2$}

\begin{equation}\label{opti-nmf}
   F( {\bf{W,H}})= \min_{ {\bf{W,H\geq 0}}}\frac{1}{2}\| {\bf{X-WH}}\|_F^2.
\end{equation}

Most of the methods for computing NMF are based on iterative update rules. A popular set of update rules given below is due  to Lee and Seung  \cite{lees}, the iteration number is in superscript. 


\begin{eqnarray}
         {\bf{W}}^{t+1}_{ia}&=&{\bf{W}}^{t}_{ia}\frac{({\bf{X}}{{\bf{H}}^{t}}^{T})_{ia}}{({\bf{W}}^{t}{\bf{H}}^{t}{{\bf{H}}^{t}}^{T})_{ia}},
         ~~\forall i,a; \label{W_Update}\\
         {\bf{H}}^{t+1}_{bj}&=&{\bf{H}}^{t}_{bj}\frac{({{\bf{W}}^{t+1}}^{T}{\bf{X}})_{bj}}{({{\bf{W}}^{t+1}}^{T}{\bf{W}}^{t+1}{\bf{H}}^{t})_{bj}},
         ~\forall b,j.\label{H_Update}
\end{eqnarray}

For many more variations on such update rules, see \cite{Neumann0}. Optimization approaches such as block-coordinate descent, projected gradient descent, and alternating non-negative least squares (ANLS) \cite{Chih-nmf-2007} have also been used for NMF.  ANLS transforms the problem in~\eqref{opti-nmf} into two convex optimization problems:

\begin{eqnarray}
{\bf{W}}_{t+1} &=& \min_{{\bf{W\geq0}}}f({\bf{W,H}}_t)  ={\min_{{\bf{W\geq0}}}\frac{1}{2}\|{\bf{X-WH}}_t\|_{F}^2,}\label{nmf-1}\\
{\bf{H}}_{t+1} &=& \min_{{\bf{H\geq0}}}f({\bf{W}}_{t+1},{\bf{H}}) =\min_{{\bf{H\geq0}}}\frac{1}{2}\|{\bf{X-W}}_{t+1}{\bf{H}}\|_{F}^2.\label{nmf-2}
\end{eqnarray}

We can solve the optimization problems given by \eqref{nmf-1} and \eqref{nmf-2} in a few ways. \cite{ab,ba} gave the Rank-one Residue Iteration (RRI) algorithm for computing NMF. This algorithm was also independently proposed  by Cichocki et al.
 \cite{CICHOCKI2}, which is called the Hierarchical Alternating Least Squares (HALS) algorithm. The solution to~(\ref{nmf-1}) and (\ref{nmf-2}) in HALS/RRI is given by explicit formulas, which make for easy implementation. Kim
 et al. \cite{DHILLON} used Newton and quasi-Newton methods to solve \eqref{nmf-1}, \eqref{nmf-2} and showed that their method has faster convergence. However, these methods require determining a suitable active set of the constraints in each iteration \cite{cvss}. Two efficient algorithms for approximately
orthogonal NMF were given by Li et al. \cite{li2014two}. Asymmetric NMF with Beta-divergences approach was studied by Lee et al. \cite{lee2009semi}.

NMF is a quadratic boolean optimization problem, so it can also be solved using the Quantum Simulated Annealing (QSA) approach of Farhi et al. \cite{farhi2000quantum}. Recently, Golden and O’Malley \cite{golden2021reverse} used a combination of forward and reverse annealing in the quantum annealing to obtain improved performance of QSA for NMF.

\subsection{Binary Orthogonal NMF}

Given a non-negative matrix ${\bf{X}}\in\Bbb{R}^{m\times n}$, a BONMF  of ${\bf{X}}$ finds the non-negative matrix ${\bf{W}}\in \Bbb{R}^{m\times k}$ and a binary ${\bf{H}}\in \{0,1\}^{k\times n}$ with $k\ll \min(m,n)$. 
The  BONMF can be written as the following optimization problem: 

\begin{equation}\label{bin-nmf}
   F({\bf{W,H}})= \min_{{\bf{W}}\in\Bbb{R}^{m\times k}~ ,{\bf{H}}\{0,1\}^{k\times n}}\frac{1}{2}\|{\bf{X-WH}}\|_F^2.
\end{equation}

Using the ANLS approach \cite{Chih-nmf-2007} we can transform (\ref{bin-nmf}) into the following sub-problems: 

\begin{eqnarray}
{\bf{W}}_{t+1} & =&{\min_{{\bf{W\geq0}}}\frac{1}{2}\|{\bf{X-WH}}_t\|_{F}^2,}\label{bnmf-1}\\
{\bf{H}}_{t+1}  &=&\min_{{\bf{H}}\in\{0,1\}}\frac{1}{2}\|{\bf{X-W}}_{t+1}{\bf{H}}\|_{F}^2.\label{bnmf-2}
\end{eqnarray}

The problem (\ref{bnmf-1}) can be solved using the update rule (\ref{W_Update}) of \cite{lees}. 
Sub-problem \eqref{bnmf-2} is solved in two different ways in the following papers. Zhang et al.
\cite{zhang2021non}   update each row of the matrix $H$  using the following strategy:

\begin{eqnarray}
 {\bf{h}}=sgn\left({\bf{X}}^T{\bf{z}}-\frac{1}{2}{\bf{Iz}}^T{\bf{z}}-{\bf{H}}'^T{\bf{W}}'^T{\bf{z}}\right),
\end{eqnarray}
where 
\begin{equation*}
sgn(x)= \left\{
\begin{array}{ll}
1,&\qquad if ~ x>0 \\
0,&\qquad otherwise,
\end{array} \right.
\end{equation*}
and ${\bf{z}}$ is the $k$-th column of ${\bf{W}}$, and ${\bf{W}}'$ is the matrix of ${\bf{W}}$ excluding ${\bf{z}}$; ${\bf{h}}^T$ is the $k$-th row of ${\bf{H}}$ and ${\bf{H}}'$ is the matrix of ${\bf{H}}$ excluding ${\bf{h}}^T$. In addition, ${\bf{I}} \in \Bbb{R}^n$ is a vector whose entries are all one. Zdunek \cite{zdunek2008data} presented another method for updating $\bf H$ under the assumption that $\bf H$ is orthogonal, which uses simulated annealing. Since they use a different approach, we don't describe it in detail.


\section{The Algorithm}
\label{sec:algo}

This section describes our approach. The method solves the optimization problems given by \eqref{bnmf-1} and \eqref{bnmf-2}.
To solve (\ref{bnmf-1}), we use the update rule given by equation \eqref{W_Update}  \cite{lees} where $\bf W$ is computed using

$$ {\bf{W}}^{t+1}_{ia} \gets  {\bf{W}}^{t}_{ia}\frac{( {\bf{X}}{{\bf{H}}^{t}}^{T})_{ia}}{( {\bf{W}}^{t} {\bf{H}}^{t}{ {\bf{H}}^{t}}^{T})_{ia}},
         \qquad\forall i,a.$$

Given $ {\bf{X, H}}$, to solve \eqref{bnmf-2} we write the problem as: 
\begin{equation}
    F( {\bf{H}})=\min_{ {\bf{H}}\in\{0,1\}^{k \times n}}\| {\bf{X-WH}}\|_F^2.
\end{equation}

Each column of the matrix $ {\bf{H}}$ is computed in two steps 
as follows:

\begin{itemize}
\item In the first step,  we calculate the angular distance between column $i$ of $ {\bf{X}}$ and  column $j$ of matrix $ {\bf{W}}$ to obtain  $ {\bf{H}}_{j,i}$.

\begin{equation}\label{updat_h}
     {\bf{H}}_{j,i}=
    \frac{\langle  {\bf{X}}_{:,i}, {\bf{W}}_{:,j}\rangle}{\| {\bf{X}}_{:,i}\|\| {\bf{W}}_{:,j}\|},
\end{equation}
where $ {\bf{X}}_{:,i}$ denotes the $i$-th column of matrix $ {\bf{X}}$ and $\langle.,.\rangle$ is the inner product.

\item 
In the second step, the maximum value (any) in each matrix column $ {\bf{H}}$ is changed to 1, and other values are changed to 0. The process can be summarized as follows:

\begin{equation}\label{n}
 {\bf{H}}_{j,i}= \left\{
\begin{array}{ll}
1,&\qquad if ~  {\bf{H}}_{j,i}=\max  {\bf{H}}_{:,i}  \\
0,&\qquad otherwise.
\end{array} \right.
\end{equation}
\end{itemize}




The pseudo-code for the method is in Algorithm \ref{alg:ge}.
These steps are executed column by column for $ {\bf{H}}.$

\begin{algorithm}[H]
\begin{algorithmic}[1]   
\State {\sc Input:}~Matrix $ {\bf{X}}\in\Bbb{R}^{m\times n}$, and $T$
\State {\sc Output:} Matrices $ {\bf{W}}\in\Bbb{R}^{m\times k}$ and $ {\bf{H}}\in \{0,1\}^{k\times n}$
\State  Initialize matrices $ {\bf{W}}$ and $\bf H$
\State {\sc While}  \mbox{$iterations < max  ~$\&$~ \neg convergence$}
\State \quad \quad $ {\bf{W}}^{t+1}_{ia} \gets  {\bf{W}}^{t}_{ia}\frac{( {\bf{X{H}}^{t}}^{T})_{ia}}{( {\bf{W}}^{t} {\bf{H}}^{t}{ {\bf{H}}^{t}}^{T})_{ia}},
         \qquad\forall i,a;$
         
\Comment Update  $ \bf W$ using (\ref{W_Update}).
\State \quad \quad $ {\bf{H}}_{j,i}^{t+1} \gets
    \frac{\langle  {\bf{X}}_{:,i}, {\bf{W}}_{:,j}^{t+1}\rangle}{\| {\bf{X}}_{:,i}\|\| {\bf{W}}_{:,j}^{t+1}\|} \quad \forall j,i$

\Comment Update $\bf H$ using  (\ref{updat_h})

\State \quad \quad
$ {\bf{H}}_{j,i}^{t+1} \gets  \left\{
\begin{array}{ll}
1,&\qquad if ~  {\bf{H}}_{j,i}^{t+1} = \max  {\bf{H}}_{:,i}^{t+1}  \\
0,&\qquad otherwise
\end{array} \right. \quad \forall j,i$

\Comment Update $\bf H$ using  \eqref{n}
\State {\sc end}
\State \textbf{return} $\bf W$ and $\bf H$
\end{algorithmic}
\caption{BONMF}
\label{alg:ge}
\end{algorithm}

\section{Empirical Evaluation} \label{sec:empirical}
This section examines three characteristics of Algorithm \ref{alg:ge}. We study the time needed for classification, the accuracy, and the time required for computing the factorization (training time) for the data sets shown in Table \ref{dataset}. The data sets are representative of the varying complexity of machine learning; some are easy (digits), some are hard (diabetes), and some have a significant number of features (ORL). These are popular datasets from the OpenML repository \cite{vanschoren2014openml}. These data sets have multiple single label classes and serve as a nice testbed for evaluating unsupervised learning algorithms, even in deep learning.
\begin{table}[H] 
\centering
\begin{tabular}{|c| c| c |c|}
\hline
Name &
\# samples & \# features&\# classes \\
\hline
ORL& 400&4,096&40\\
\hline
Optdigits&5,619&65&10\\
\hline
Phishing&11,055&68&2\\
\hline
Monkey&471&6&2\\
\hline
Pendigits& 10,992&17&2\\
\hline 
Diabetes&7,67&8&2\\\hline
W8a&49,748&300&2\\\hline
Banking&8237&13&3\\\hline
Svmguide&3,087&5&2\\\hline
\end{tabular}
    \caption{Data Sets}
    \label{dataset}
\end{table}
We compare the performance of Algorithm \ref{alg:ge} with the algorithms for orthogonal matrix factorization \cite{ding}, non-negative matrix factorization \cite{lees}, and semi-binary non negative matrix factorization \cite{zhang2021non}. We examine the relative performance of these algorithms for accuracy and classification time. We use two methods for ONMF to classify a new data point $j$ (column vector of $\bf X$). Typically, ONMF uses the index $i$ of the maximum entry in column $ {\bf{H}}_{:,j}$ for classification, which gives the cluster to which data $j$ belongs. The data points in cluster $i$ may have different labels, and the label of $j$ is the label of the point in cluster $i$ that is closest (distance-wise). We refer to this default scheme for determining the label as ONMF in Table \ref{experiments}. The second scheme we use to determine the label uses the label on $i'$, which is the point in cluster $i$ that forms the smallest angle data point (vector for  $j$); then, the label of $i'$ is used to classify point $j$. The cluster to which data point $j$ belongs is again computed based on the angles to the columns of $\bf W$, the closest column of $\bf W$ determines the cluster, and the closest point in the cluster (angle-wise) determines the label. The second scheme is ONMF-cos in Table \ref{experiments}. The other two algorithms that we used to compare are i) the popular and the foundational algorithm of \cite{lees} for NMF, labelled "Lee and Seung" and ii) the algorithm of \cite{zhang2021non} for NMF with the constraint that the entries in $\bf H$ are binary (labelled ``Zhang et al."). We use only the matrix-based method factorization algorithms closest to the K-means for evaluation. As part of a future study, it would be interesting to see how these algorithms perform against a highly optimized implementation of K-means.

%

We report on experiments that were run on a laptop (i5–7200U, 12GB of RAM). The  algorithms \cite{lees,zhang2021non}
 were coded in Python 3.1. 
We used the number of classes as the rank in factorization. Eighty percent of the data was used for training, and the remaining was used for testing the accuracy.
We use the python library  (ionmf.factorization.onmf) for ONMF \cite{stravzar}. Initialization of $\bf W, H$ is done using the following scheme: we sort the columns of the matrix $\bf X$ based on its norm. To determine the $i^{th}$ column of $\bf W$, use the average of ten randomly chosen columns from the first thirty columns of $\bf X$ as in  \cite{albright2006algorithms}. Initial matrix $ {\bf{H}}^0$ is computed using $ {\bf{H}}^0=( {\bf{W}}^T {\bf{W}})^{-1} {\bf{W}}^T {\bf{X}}.$ Since the initial values of $\bf W$ are random, we run the algorithm thirty times and report the averages in Table \ref{experiments}. The first thing to note is that in Algorithm \ref{alg:ge} extra computation is needed to convert $\bf H$ to binary in each iteration. This computation increases the time needed for factorization relative to ONMF and NMF and is linear in the size of $\bf H$. However, given the factorization, the classification phase is more efficient, and $\bf H$ is sparse.

\subsection{Classification} \label{sec:classification}

In the basic NMF approach given by update rules (\ref{W_Update}) and (\ref{H_Update}) (as in \cite{lees}), the number of steps needed for the classification of new data (column vector of $\bf X$) is proportional to the number of columns in the factorization $\bf W, H.$ 
{We need to calculate the angle  between  the coefficient vector for the new data and all
} the columns of $\bf H$ (as many as the columns in $\bf X$) to determine the label for the data. 
Algorithm \ref{alg:ge} does not share this disadvantage. We can compute the angle of the sample to every column of $\bf W$ (a low-rank matrix) and use the closest column to determine the label. This observation is reflected in data in the row labelled ``TT  (s)" in Table \ref{experiments}.

\subsection{Accuracy} 

The accuracy of the five methods is presented in Table \ref{experiments}. The entries in bold font indicate that a particular method had the most accuracy). Six of eight data sets (except pendigits and banking) have improved accuracy for classification when $\cos$ angles are used to measure similarity, and the utility of using the angle (\ref{updat_h}) is evident. The method presented here has the best accuracy on six of the eight data sets (expect optdigits, phishing). Regarding classification time, it performs best on seven of the eight datasets. Note that the other method ONMF+$\cos$ which is as competitive as Algorithm \ref{alg:ge} uses $O(nk)$ space to store $\bf H$ whereas Algorithm \ref{alg:ge} uses $O(n)$ space even in the intermediate stages of the calculations.

\begin{table*}
    \centering
\begin{tabular}{|c|c| c| c |c|c|c|}
\hline
Name& & ONMF 
& 
& Algorithm \ref{alg:ge} 
& ONMF+ $\cos$  
&  
\\
& & Strazar et al. 
& Lee and Seung 
&  
&Strazar et al.
& Zhang et al.
\\
\hline
ORL&TT (s)&\bf 1.62&1.63&2.28& 1.63&9.48\\
& CT (s)&0.24&0.178&{\bf 0.09} &{\bf 0.09}&0.20\\
& AC (\%)&85.00&85.00&{\bf 89.99} &{\bf 89.99} &85.00 \\
\hline
Optdigits&TT  (s)&0.38&0.62&2.60&\bf 0.36&3.72\\
&CT  (s)&45.24&38.97&\bf13.03&13.23&37.91\\
& AC (\%)&53.45 &62.27&80.78&\bf 88.96& 48.12\\
\hline
Phishing&TT  (s)&0.87&\bf 0.80&3.42&1.05&1.96\\
&CT  (s)&156.24&142.92&\bf 52.22&  53.14&146.40\\
& AC (\%)&54.76&54.76&91.85&\bf 92.14&54.76\\
\hline
Monkey&TT  (s)&\bf 0.01&\bf 0.01&0.10&\bf 0.01&0.06\\
&CT  (s)&0.45&0.34& 0.14&\bf 0.12&0.31\\
& AC (\%)&48.80&48.80&\bf 80.95&\bf 80.95&53.12\\
\hline
Diabetes&TT  (s)&0.02&\bf0.01&1.016&\bf 0.01&0.59\\
&CT  (s)&0.80&0.85&\bf 0.24& 0.25&0.67\\
& AC (\%)&51.72&51.72&\bf 68.96&\bf 68.96&68.95\\
\hline
Banking&TT  (s)&\bf0.05&0.06& 2.01& 0.07&1.56\\
&CT  (s)&77.07&70.87&\bf 25.25&27.42&73.74\\
& AC (\%)&\bf 87.05& 76.09&\bf 87.05&\bf 87.05&\bf 87.05\\
\hline
Svmguide&TT  (s)&\bf 0.01&\bf0.01&0.82&0.02&0.40\\
&CT  (s)&13.90&12.05&\bf 4.93&45.27&13.51\\
& AC (\%)&73.70&65.73&\bf 80.17&\bf 80.17&79.23\\
\hline
Pendigits&TT  (s)&2.33&2.72& 5.08&\bf 2.22&4.86\\
&CT  (s)&161.30&161.50&\bf 53.19& 54.39&170.83\\
& AC (\%)&\bf90.47&82.29&\bf90.47&\bf \bf90.47&\bf90.47\\
\hline
W8a&TT  (s)&29.80&27.25& 37.96&\bf 26.30&32.93\\
&CT  (s)&2851.52&2721.82&\bf 999.72& 1006.71&2659.24\\
& AC (\%)&95.25&95.25&\bf97.14&\bf \bf97.14&\bf97.14\\
\hline
\end{tabular}
\begin{itemize}
        \item[1.] TT is the training time in seconds. 
        \item[2.] CT is the classification time in seconds. 
        \item[3.] AC is the accuracy in $\%$. 
\end{itemize}
\caption{Numerical Results}
\label{experiments}   
\end{table*}




\subsection{Running Time and Space}

Table \ref{experiments} shows the running time for the training and classification phase of the five algorithms. The entries in bold signify that the running time is the smallest. The proposed algorithm \ref{alg:ge} is as fast and accurate as the other best method in the table, which is the modified version of ONMF in which angles are used for classification. However, we cannot directly compare the factorization returned by the two algorithms, as ONMF returns $\bf H$ with orthogonal rows in which entries are real. In contrast, the method proposed here returns a sparse $\bf H$, which is binary and has orthogonal rows. Algorithm \ref{alg:ge} is space efficient compared to the ONMF+cos method, which is an essential consideration for large data sets.

Based on the data in Table \ref{experiments}, we can conclude that Algorithm \ref{alg:ge} has competitive accuracy and leads to better clustering with a natural interpretation and a sparse representation. It also classifies new data faster.

 
\section{Conclusion}

This paper gives a new geometric approach for binary orthogonal non-negative matrix factorization. The proposed method is space efficient.
We also compared the proposed Algorithm with three other methods \cite{lees,zhang2021non,ding} for accuracy and time on representative datasets in machine learning. Our experiments show that the method is fast and accurate on the data sets tested.
\section*{Acknowledgements}
The authors would like to thank Chirag Wadhwa for discussion and comments.

\end{document}